\documentclass[11pt]{article}
\usepackage[preprint]{acl}   

\usepackage{times}

\usepackage{microtype}
\usepackage{graphicx}
\usepackage{subfigure}
\usepackage{booktabs} 
\usepackage{hyperref}

\usepackage{tabularx}
\usepackage{tabulary}
\usepackage{multirow}
\usepackage{longtable}

\usepackage{amsmath}
\usepackage{amssymb}
\usepackage{mathtools}
\usepackage{amsthm}

\usepackage{colortbl}

\newcommand{\SA}{$SA$ }
\newcommand{\SAN}{$SA_{n}$ }

\newcommand{\SAC}{$SA$, }
\newcommand{\SC}{$SC$ }

\newcommand{\SCC}{$SC$, }

\newcommand{\ST}{$ST$ }
\newcommand{\STC}{$ST$, }

\newcommand{\SV}{$SV$ }
\newcommand{\MD}{$MD$ }

\newcommand{\MT}{$MT$ }

\newcommand{\ND}{$ND$ }

\newcommand{\OPR}{$OPR$ }
\newcommand{\SGN}{$SGN$ }

\newcommand{\DN}{$D_{n}$}
\newcommand{\DND}{$D'_{n}$}
\newcommand{\ANST}{$ST_{n}$ }

\newcommand{\AMSV}{$SV_{n-1}$ }

\newcommand{\AN}{$A_{n}$}

\definecolor{lightbrown}{rgb}{0.7, 0.4, 0.1}

\title{Understanding Addition and Subtraction in Transformers}

\author{%
    Philip Quirke\\
    Martian\\
    \texttt{philip@withmartian.com}
    \And
    Clement Neo\\
    NTU\\
    \And
    Fazl Barez\\
    University of Oxford, Martian\\
    \texttt{fazl@robots.ox.ac.uk}
}

\begin{document}
\maketitle

\begin{abstract}

We use integer addition and subtraction as a controlled, exactly-solvable testbed for what can be said with confidence about
the algorithm a low-loss transformer implements — logically and mechanically. We train small transformers (2–3 layers) from
scratch, find the edge cases they fail (long carry and borrow cascades), and enrich the training data with them; most
resulting models reach $>$99.999\% accuracy on 5–15 digit problems in under an hour, solving cascades such as
555555555+444444448=+1000000003. Interpretability analysis of these accurate models — not an a-priori guess — surfaces the
structure from which we derive novel, sound, exact left-to-right algorithms for both operations, matching the causal ordering
transformers must use. We then put these algorithms back to the models: across 46 models varying in seed, depth, digit-length,
and operation, the accurate ones share one circuit skeleton, and in representative models deeper causal analysis (activation
and edge path-patching, with controls) confirms the multi-digit resolved-carry sub-task that earlier analyses left open —
while ruling out several candidate mechanisms. The skeleton recurs across model sizes and operations, and mixed-model nodes
are polysemantic across addition and subtraction whether initialized from an addition model or trained from scratch. We
release the models, analysis, and a reproducible interpretability toolkit.

\end{abstract} 

\section{Introduction}

\begin{figure}[h]
\centering
\includegraphics[width=\columnwidth]{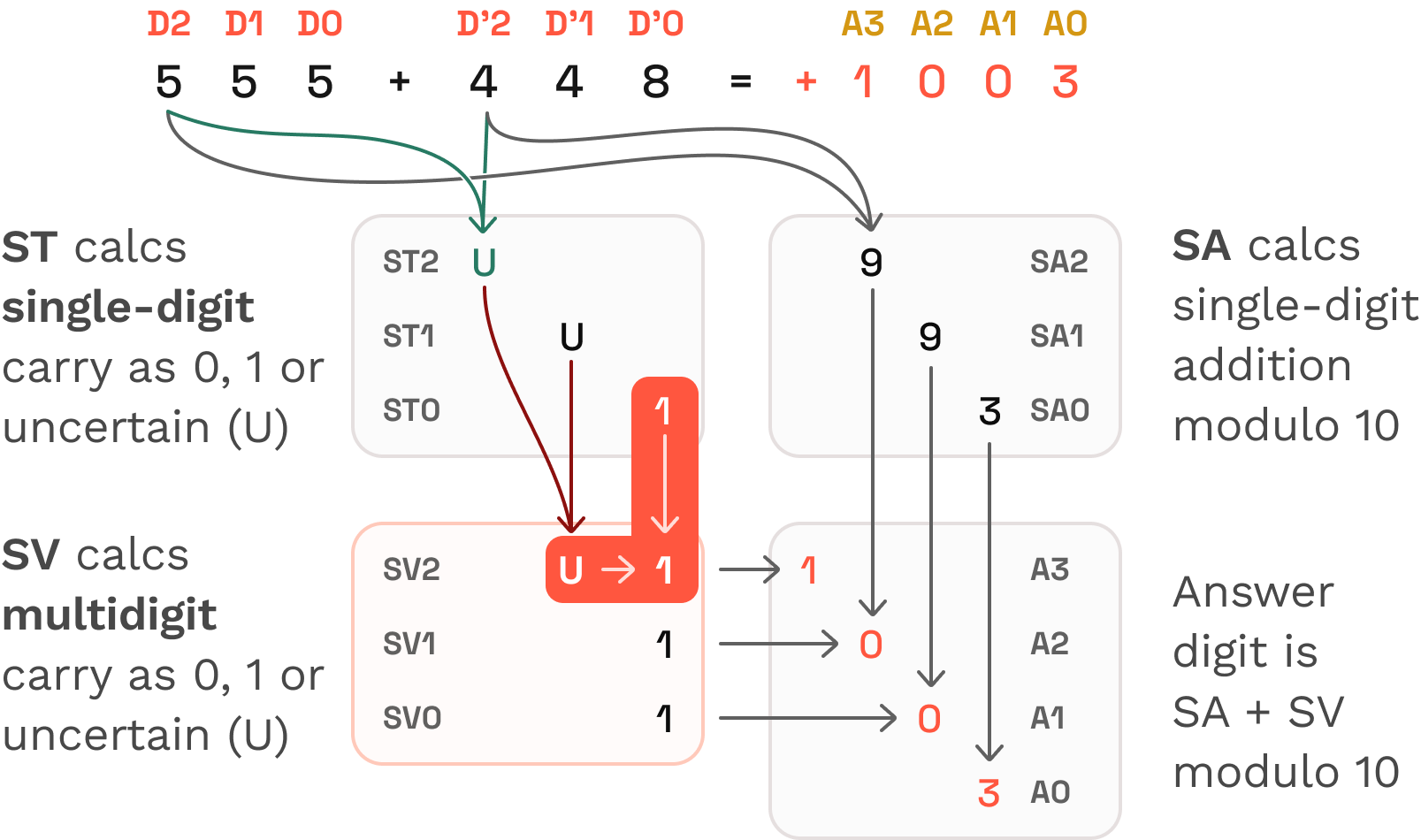}
\caption{Our novel n-digit addition algorithm is sound. Working left to right, it calculates single-digit carry-one values $(ST_n$),
combining them into multidigit carry-one values $(SV_n$). Uncertain $SV$ values (U) are progressively refined to 0 or 1 (highlighted)
until by the “=" token there is no $SV$ uncertainty. The answer digits \AN $ $ are calculated from $SA_n$ and \AMSV values.}
\label{fig:AdditionAlgorithm}
\end{figure}

Mechanistic interpretability seeks to recover the algorithm a trained network implements, but progress is hard to judge without ground
truth: for most tasks the ``correct'' algorithm is unknown, so a proposed circuit may be genuine or merely a plausible-looking artifact.
Integer addition and subtraction give us more to work with. The correct output for any input is computable, and we understand where the
difficulty lies---the long carry and borrow cascades that are the hardest edge cases. Knowing at least one exact algorithm, and where it is
stressed, lets us design targeted, falsifiable tests of a proposed mechanism---even though a trained model may implement a different valid
algorithm than one we would write by hand. We use this as a controlled testbed for what can be said \emph{with confidence} about a low-loss
transformer's algorithm---logically and mechanically---and we treat our own account as a hypothesis to be attacked, not defended.

We did not begin with an algorithm. We first trained small transformers (2--3 layers, 3--4 heads) on 5- to 15-digit addition, subtraction,
or both, examined the edge cases early models failed---long carry and borrow cascades---and enriched the effectively-infinite synthetic
training data to remove them. Most of these models converge in under an hour on a single GPU to $>$99.999\% accuracy, solving cascades such
as $555555555+444444448=+1000000003$, and at $\sim$10M parameters form an efficient, reproducible, interpretable testbed
(Sec.~\ref{sec:TrainingModels}). The problem space is astronomical---the 5-digit task alone has $10^{10}$ distinct
questions, against $\sim$10M parameters---and each training step draws fresh data, so memorization is
infeasible: to reach this accuracy a model must compute the answer (App.~\ref{app:TrainingData}).

We then analyzed these near-perfect models. Attention patterns, ablations, and PCA of node outputs surfaced structure recurring
across models; the algorithms we present (Fig.~\ref{fig:AdditionAlgorithm} and Fig.~\ref{fig:SubtractionAlgorithm}) are our attempt to explain
that structure, not a template imposed on it. They are novel, sound, and exact, and (unlike humans) process digits left-to-right---the order
in which arithmetic appears in text and under which LLMs do inference.

We then subject this derived account to independent causal tests designed so it can fail. Earlier candidate accounts that did not survive
our controls were discarded.
Our tests proceed in two tiers. First, an automated search over every model finds candidate ``nodes'' for each sub-task and verifies, by
ablation, that they satisfy the algorithm's attention, position, and ordering constraints (our 6-digit model alone has over 30). Every
accurate model contains the same circuit skeleton, though the assignment of sub-tasks to heads---and the degree of redundancy---varies
across models. Second, on four representative models we use activation and edge path-patching to causally confirm the multi-digit resolved
carry ($SV$)---the sub-task earlier analyses left open---tracing how carries are computed, propagated, resolved, and combined into the answer.

To make this analysis reproducible, we release our interpretability library (which
provides node characterization, search, and ablation; testing of node relationships and constraints; and visualizations),
training data and models.\footnote{Code: \href{https://github.com/PhilipQuirke/quanta_maths}{github.com/PhilipQuirke/quanta\_maths}.
Models (Hugging Face collections):
\href{https://huggingface.co/collections/PhilipQuirke/quantamaths-addition-models}{addition},
\href{https://huggingface.co/collections/PhilipQuirke/quantamaths-subtraction-models}{subtraction},
\href{https://huggingface.co/collections/PhilipQuirke/quantamaths-mixed-models}{mixed}.}

\textbf{Our contributions are fivefold}: (i) enriched datasets for n-digit addition and subtraction; (ii) 46 small transformers trained from 
scratch--- many to $>$99.999\% accuracy; (iii) novel, sound, exact, human-comprehensible left-to-right addition and subtraction algorithms; (iv) causal 
evidence---across a family of models, and closing the prior gap on the multi-digit resolved-carry sub-task---that these models implement the 
algorithms; and (v) a reusable interpretability toolkit.

We conclude that transformers can implement exact n-digit addition and subtraction through interpretable circuits. 
With appropriate training and data, larger LLMs could in principle adopt the same mechanisms.

\section{Related Work}
\label{sec:RelatedWork}

\textbf{Mechanistic understanding of arithmetic in transformers}. Early work by \citet{nanda2023progress} revealed that one-layer models
implement modular addition through discrete Fourier transforms, converting addition into rotations in frequency space---showing transformers
can find unintuitive algorithms for basic mathematics. \citet{quirke2023understanding} detailed an algorithm for n-digit addition in a
single-layer model that achieved 99\% accuracy, identifying a cascading carry-one failure mode. Subsequent work on training dynamics
\citep{musat2024clusteringalignmentunderstandingtraining} and the finding that even random transformers exhibit algorithmic capabilities
\citep{zhong2024algorithmiccapabilitiesrandomtransformers} suggest architectural biases toward systematic computation. Whether arithmetic
models memorize or generalize is central \citep{mahdavi2024memorization,michaud2023quantization}; our large problem space, small models, and
fresh-data training place these models firmly in the generalizing regime.

\textbf{Architectural modifications and specialized models}. Near-perfect arithmetic accuracy is achievable through architectural
changes: least-significant-digit-first inputs with position embeddings and zero-padding
\citep{mcleish2024transformersarithmeticrightembeddings}, or aligning computational flow with carry propagation
\citep{cho2024positioncouplingimprovinglength}. Specialized models excel: MathGLM's 10M parameter model achieves 100\% accuracy on
integer addition through step-by-step reasoning \citep{yang2023gptsolvemathematicalproblems}, while
\citet{qiu2024dissectingmultiplicationtransformersinsights} reached 99.9\% on 5-digit multiplication with a tiny transformer that
outperforms GPT-4. Subtraction is much less studied, though \cite{zhang2024interpretingimprovinglargelanguage} identified symmetries between
addition and subtraction circuits, with the same attention heads handling both operations.

\textbf{Interpretability methods}. Our analysis uses standard circuit-analysis tools: ablation and activation patching (interchange
interventions) to test a node's causal role \citep{FactualAssociations,GoldowskyDill2023LocalizingMB}, edge/path patching and causal
scrubbing to test connections between nodes \citep{causalscrubbing2023}, the logit lens \citep{logitlens2020}, and sparse autoencoders
\citep{cunningham2023sparse}, within the transformer-circuits framework \citep{elhage2021mathematical}. Our automated search for
role-specific nodes is close in spirit to automated circuit discovery \citep{Conmy2023TowardsAC}; see \citet{bereska2024mechanistic} for a
survey.

\textbf{Production systems versus specialized models}. The Claude 3.5 Haiku circuit analysis \cite{lindsey2025biology} reveals how frontier
models implement addition differently: using ``bespoke" circuits combining lookup tables with magnitude estimation rather than clean n-digit
algorithms. This illustrates how production LLMs develop hybrid strategies instead of the interpretable, generalizable algorithms that
specialized models achieve.

\section{Methodology}
The algorithms below were derived from the analysis of Sec.~\ref{Sub:ExperimentalResults}; we present them first for clarity. Transformer
models can learn algorithms different from traditional human methods. We define an alternative, mathematically-equivalent framework for
addition and subtraction here and later sections demonstrate our models implement these approaches.

\subsection{Mathematical Framework}

\begin{figure}[h]
\centering
\includegraphics[width=\columnwidth]{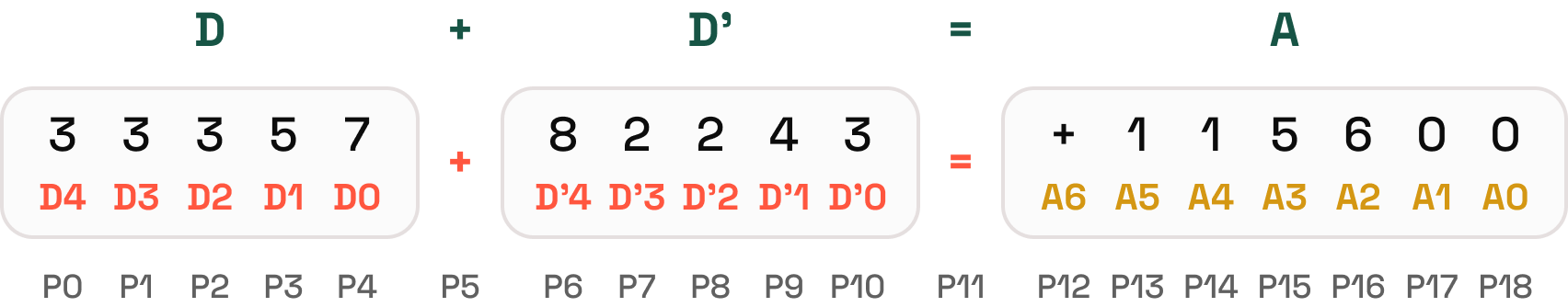}
\caption{For 5-digit addition and subtraction, our notation for the main input tokens is $D4$, ..., $D0$ and $D'4$, ..., $D'0$. 
For output tokens it is $A6$, .., $A0$. For n-digits, we use the notation \DN, .., $D0$, \DND, .., $D'0$ and \AN, .., $A0$.}
\label{fig:AdditionTerms}
\end{figure}

For addition and subtraction of two $n$-digit numbers, we use the Fig.~\ref{fig:AdditionTerms} notation. (Detail in App.~\ref{app:Terminology}). 

\subsection{Addition Algorithm Description}
\label{Sec:AdditionFramework}

For addition, aligned with \citet{quirke2023understanding}, we define the “Base Add" subtask $SA_n$ to compute the digit-wise sum modulo 10:

\begin{equation}
    SA_n = (D_n + D'_n) \mod 10  
\end{equation}

Following \citeauthor{quirke2023understanding}, the binary ``Simple Carry'' subtask $SC_n$ records whether a digit pair alone produces a carry:

\begin{equation}
    SC_n = 1 \text{ if } D_n + D'_n \geq 10 \text{ else } 0
\end{equation}

Using only $SA_n$ and $SC_n$, their 1-layer model predicts single-digit carries (e.g.\ ``045+046=0091'')
but fails when a carry cascades through two or more digits (e.g.\ ``555+446'' predicted as ``0901''). The tri-state $ST_n$ below supersedes
$SC_n$; we nonetheless find some models still compute the logically redundant $SC_n$.

To solve $n$-digit addition with high accuracy, the framework must handle carry bits that cascade through multiple digits: to predict
the first answer digit of “555+448=+1003" as “1", the carry bit must cascade from the \textit{rightmost} to the \textit{leftmost} digit
in a single forward pass---challenging for an autoregressive model that processes tokens left to right. We define two new subtasks for this.

First, we introduce subtask ST. It classifies a digit pair sum as \textit{definitely} causing a carry (e.g. 6+7), definitely \textit{not} 
causing a carry (e.g. 2+3), or \textit{possibly} causing a carry (e.g. 5+4). The 5+4 case is uncertain as the digits sum to 9 and a carry 
from the next-lower-value digit would cause a carry:

\begin{equation}
    \underbrace{ST_n}_{(D_n, D'_n)} =
    \begin{cases}
    0 & \text{if } (D_n + D'_n) \le 8 \text{ or } n = 0 \\
    1 & \text{if } (D_n + D'_n) \geq 10 \\
    U & \text{otherwise. (U for uncertain)} \\
    \end{cases}
\end{equation}

The “TriAdd" function combines two $ST_n$ style values X and Y. Only if \textit{both} X and Y are U, does TriAdd return U (there is still uncertainty), 
otherwise it returns 0 or 1 (the carry bit value is known): 

\begin{equation}
    \underbrace{\text{TriAdd}}_{(X, Y)} = \\
    \begin{cases}
        Y & \text{if } X = U \\
        X & \text{otherwise} \\
    \end{cases}
\end{equation}

\begin{figure*}[t]
\centering
\includegraphics[width=0.9\textwidth]{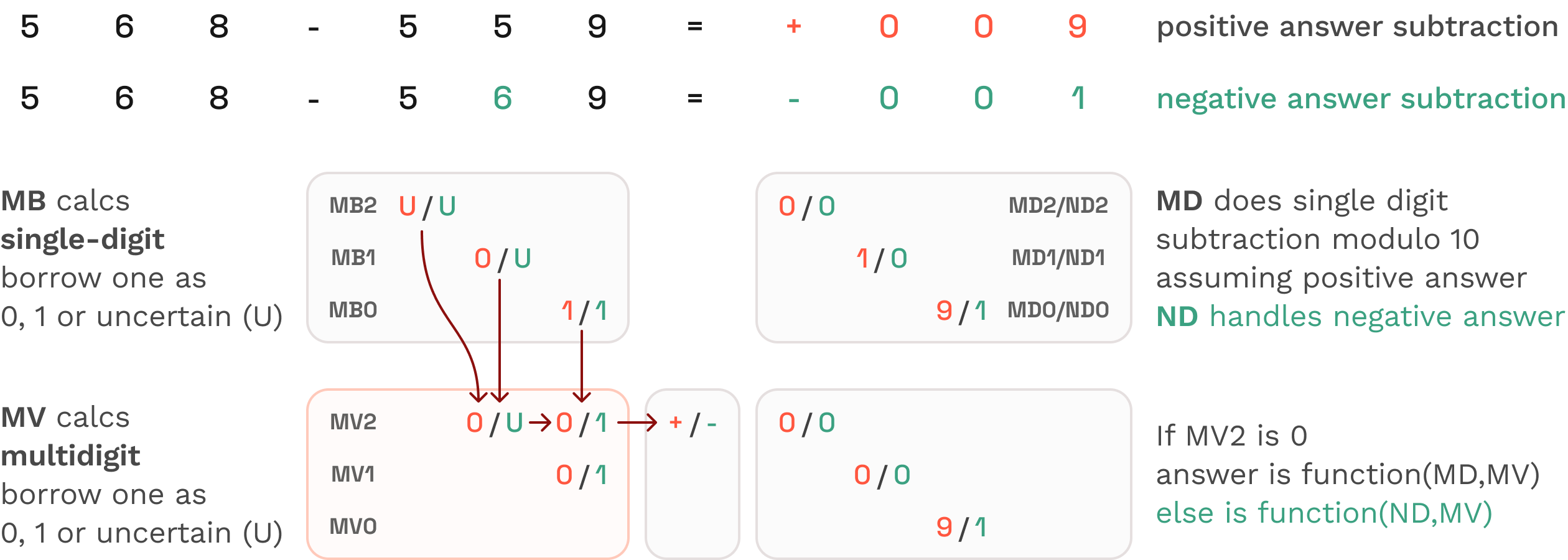}
\caption{
Our novel n-digit subtraction algorithm is sound. It parallels our addition algorithm but instead refines “cascading borrow one" uncertainty over multiple tokens. 
The refined $MV2$ value determines both the answer sign (“+" or “-") and which digit values ($MD_n$ or $ND_n$) to use for the final output.}
\label{fig:SubtractionAlgorithm}
\end{figure*}

Finally, we introduce the SV subtask that uses TriAdd to handle multidigit carry cascades. We describe $SV_n$ using the first 3 examples:

\begin{align}
    SV0 &= ST0 \\
    SV1 &= \text{TriAdd}(ST1,ST0) \\
    SV2 &= \text{TriAdd}(\text{TriAdd}(ST2,ST1),ST0)
\end{align}

The highlighted (solid-color) part of Fig.~\ref{fig:AdditionAlgorithm} illustrates SV2: ST2 and ST1 (the higher-value digits) combine first,
mirroring the left-to-right order in which the model processes tokens. The last term, ST0, is always 0 or 1, so every $SV_n$ evaluates
to 0 or 1. That is by the end of the calculation all uncertainty is resolved.

The $SV_n$ are accurate multidigit cascade carry values, computed across multiple tokens and complete by the “=" token; combined with
the \SAN values (Fig.~\ref{fig:AdditionAlgorithm}) they give accurate answer tokens.

\subsection{Subtraction Algorithm Description}
\label{Sec:SubtractionFramework}

Subtraction has been studied far less than addition (Sec.~\ref{sec:RelatedWork}). We give a novel, exact left-to-right algorithm for it that
parallels our addition algorithm, replacing the “cascading carry one" with a “cascading borrow one" and reducing the same kind of
token-by-token uncertainty.

We introduce “Base Diff" $MD_n$ defined as \DN - \DND $ $ modulo 10 (paralleling $SA_n$). We also introduce $MB_n$ for the single-digit “borrow one" case (paralleling $ST_n$):

\begin{equation}
    \underbrace{MB_n}_{(D_n, D'_n)} = \\
    \begin{cases}
        1 & \text{if } D_n < D'_n  \text{ (borrow one)}\\
        0 & \text{if } D_n > D'_n \text{ (no borrow)}\\
        U & \text{if } D_n = D'_n \text{ (uncertain)} \\
    \end{cases}
\end{equation}

Finally, we introduce $MV_n$ which parallels $SV_n$ to handle the “cascading borrow one" edge case. With this formulation, the addition and subtraction algorithms 
have the same structure (just replacing $SA_n$ with $MD_n$, $ST_n$ with $MB_n$, and $SV_n$ with $MV_n$). 

Subtraction poses an additional difficulty: some questions give a positive answer (e.g. 325-129=+196) and others a negative one
(e.g. 325-329=-004), with every answer digit differing between the two. Guided by experimental insights, we introduce “Neg Diff" $ND_n$
defined as \DND - \DN $ $ modulo 10 (the opposite of $MD_n$). At the “=" token the $MV_n$ values are known, and the algorithm
selects $MD_n$ or $ND_n$ as appropriate.

In Fig.~\ref{fig:SubtractionAlgorithm}, the value $MV2$ determines the answer sign. While the following calculations could use $MV2$ too, 
in fact they attend to the answer sign itself. We introduce the subtask SGN to reflect this.     

\subsection{Mixed Algorithm Description}
\label{Sec:MixedFramework}

For mixed models handling both operations, the addition and subtraction sub-tasks above still apply, but the model must also decide, per
question, which operation's answer to emit. We add the subtask \OPR (“attends to the prompt operation token"). Using OPR and the answer sign
SGN, a mixed model selects between addition ($SA_n$), positive-answer subtraction ($MD_n$), and negative-answer subtraction ($ND_n$)
outputs. Sec.~\ref{Sub:ExperimentalResults} shows that individual nodes become polysemantic, computing several of these in parallel.

\section{Experiments}

\begin{figure*}[t]
\centering
\includegraphics[width=0.9\textwidth]{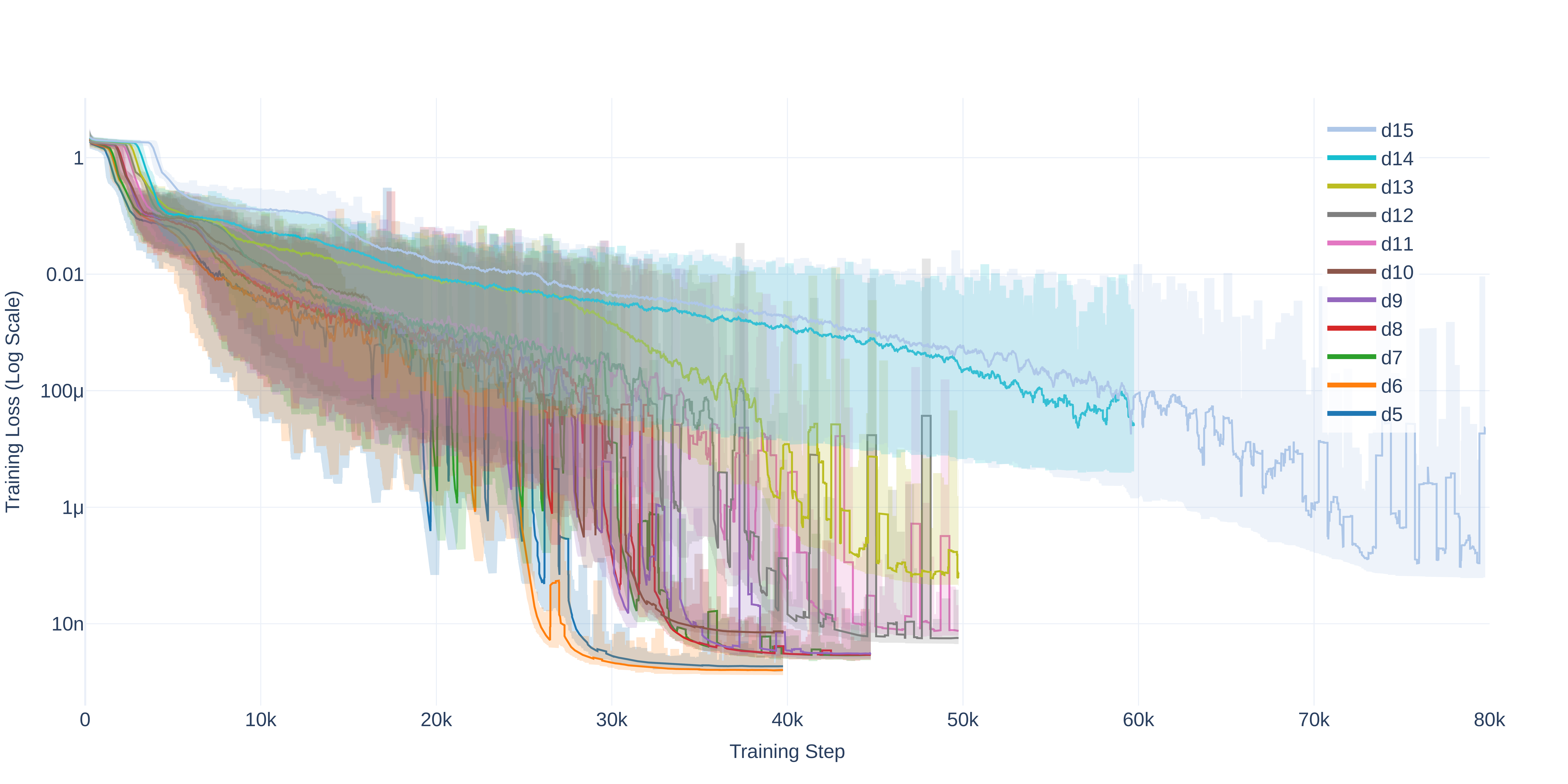}
\caption{The 5- to 15-digit 2-layer 3-head addition models have very low loss. With more digits training takes longer. Details in Tab.\ref{tab:AdditionModels}.}
\label{fig:TrainingLossAddition}
\end{figure*}
\subsection{Training Models}  
\label{sec:TrainingModels}

Our models' vocabulary consisted of 14 tokens: the digits 0-9 plus the mathematical operators +, -, =, *, and /.
We created an infinite training dataset enriched with rare numerical combinations (mainly cascading use cases). (Also see App.~\ref{app:TrainingData})

We trained addition-only, subtraction-only, and mixed (addition and subtraction) models. We systematically explored different model architectures
and settled on 2--3 layers (2 for addition and subtraction, 3 for mixed) and 3--4 attention heads. (Also see App.~\ref{app:ModelShape})

We used a batch size of 64, learning rate of 0.00008, and weight decay of 0.1. The loss function was defined as the mean negative log likelihood 
across all output tokens. Training was stopped when the loss was very low e.g. $2 \times 10^{-8}$. (Also see App.~\ref{app:ModelLoss}, Fig.~\ref{fig:TrainingLossAddition}, Fig.~\ref{fig:TrainingLossMixed}). 

Each model was tested on 1 million questions; for ease of presentation we say a model has 99.999\% accuracy if fewer than 10 were wrong
(Clopper-Pearson intervals in the tables). Most models achieved this. The 46 models are summarized in Tab.~\ref{tab:AdditionModels} and
Tab.~\ref{tab:MixedModels}.

\begin{figure}[ht]
\centering
\includegraphics[width=\columnwidth]{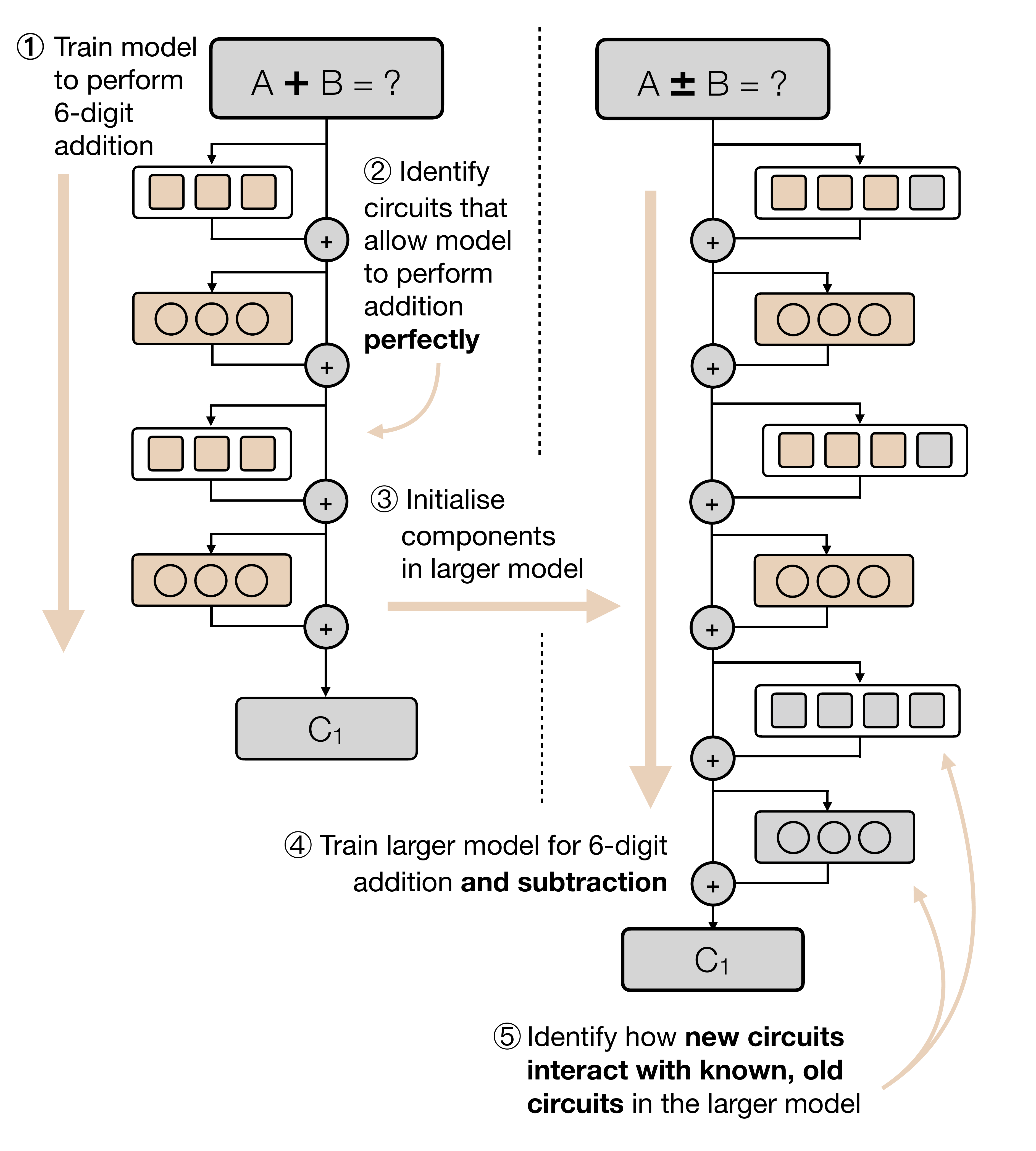}
\caption{Mixed model initialization and analysis: (1) Train an accurate 5-digit addition model. (2) Reverse-engineer it to identify the
implemented algorithm. (3) Insert its attention head and MLP weights (in brown) into a larger model. (4) Train the new model on 80\%
subtraction and 20\% addition. (5) The resulting model predicts accurately and reuses the inserted addition circuits for both tasks.}
\label{fig:InsertAddIntoMix} 
\end{figure}

\subsubsection{Training mixed models}
\label{sec:TrainingNDigitMixed}

To explore how transformers learn multiple operations simultaneously---and whether addition-specialized knowledge transfers---we trained
models capable of both addition and subtraction.

We trained mixed models in two ways: 9 from scratch, and 14 initialized from a highly accurate addition-only model to test whether addition
knowledge transfers (5- to 13-digit, 2--4 layers, 3--4 heads; see Tab.~\ref{tab:MixedModels}). For the initialized models we copied the addition
weights into the first two layers and first three heads, then trained with a curriculum of 80\% subtraction and 20\% addition using enriched
data for both operations (Fig.~\ref{fig:InsertAddIntoMix}). From-scratch mixed models also train to high accuracy; the initialization
experiments test transfer specifically.

Initialization proved effective: most models quickly reached high accuracy on both operations. Periodically ``freezing'' the inserted
heads or MLPs (recopying their addition weights every 100 steps) lowered accuracy---the initialized weights need freedom to adapt
(App.~\ref{app:MixedModelInit}).

A seed-sensitivity analysis appears in App.~\ref{app:SeedSensitivity} (Fig.~\ref{fig:SeedSensitivity}).

\subsection{Experimental Results}
\label{Sub:ExperimentalResults}

To test whether our models \emph{implement} the addition, subtraction and mixed algorithms above---rather than merely correlating with them---we 
analyze their internals in two tiers: (i) an automated search across the whole family for the circuit skeleton the algorithms require, and (ii) 
a deep causal study, on four worked-example models (6- and 8-digit addition; 6- and 8-digit mixed), of the sub-task that prior work left 
unconfirmed: the multi-digit resolved carry (\SV).

\subsubsection{Shared circuit skeleton across the family}
\label{sec:SkeletonFamily}

\begin{figure}[h]
\centering
\includegraphics[width=\columnwidth]{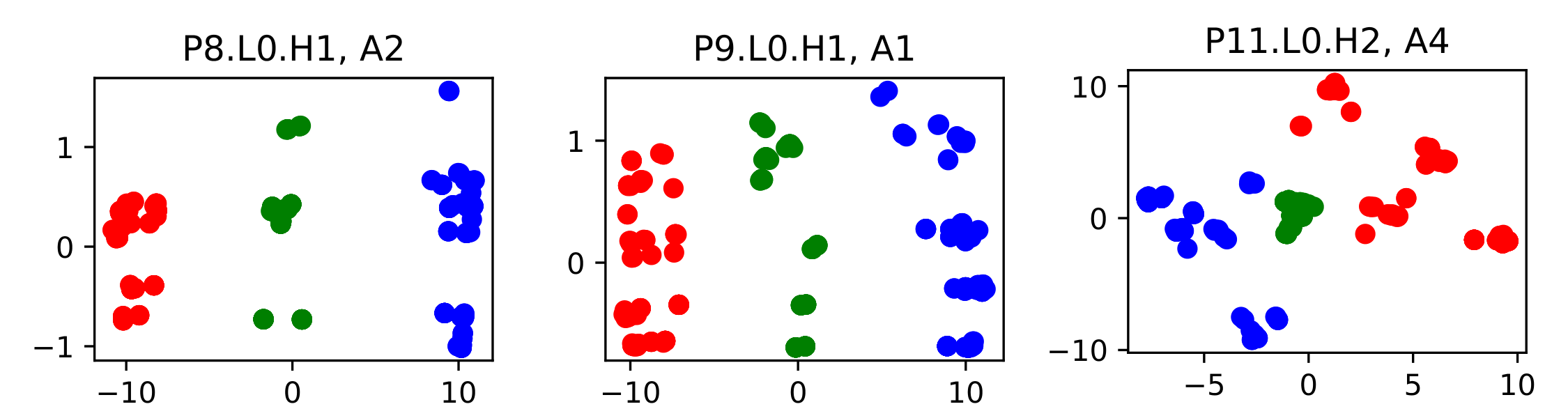}
\caption{For a 5-digit addition model, these three sample PCAs each show 1 attention head at a token position tested for 1 answer digit. 
Each dot represents a question, colored by the question's expected ST value (0, 1, or U). The three ST classes form distinct clusters in each plot. 
Cluster geometry varies across heads and models, but the three-way separation is always present; Sec.~\ref{sec:CausalSV} tests these writes causally.}
\label{fig:PCATrigrams5D}
\end{figure}

For each model we search every (sub-task, digit-position) pair for candidate nodes, keeping only those that satisfy the algorithm's constraints:
the node attends to the relevant operand tokens, occupies a valid position, exhibits the expected sub-task structure (e.g.\ three $ST$ clusters
under PCA; Fig.~\ref{fig:PCATrigrams5D}), and respects the ordering dependencies (e.g.\ $SV_1 = \mathrm{TriAdd}(ST_1, ST_0)$ must follow its inputs---our 
6-digit model has over 30 such constraints). We then ablate each candidate to confirm its predicted effect (full constraint list and
techniques: App.~\ref{app:Hypothesis3}, App.~\ref{app:Techniques}).

The 1-layer baseline of \citet{quirke2023understanding} reaches only 99\% accuracy using per-digit \SAC \SC nodes. Of our 46 trained
models, 29 meet the fewer-than-10-failures-per-million bar (Tabs.~\ref{tab:AdditionModels}, \ref{tab:MixedModels}), and in all 27 accurate
addition and mixed models the search finds the \emph{same} skeleton: every required sub-task present at a valid position, every ordering
constraint holding. This is a per-model, machine-checked result: the underlying per-model maps are released, making the family claim
independently reproducible. Map \emph{depth} varies with size (the largest maps lack some consumer-head tags); Tab.~\ref{tab:Techniques}
records which technique ran at which scale.
Subtraction-only training mostly falls short of the accuracy bar (Tab.~\ref{tab:MixedModels}); in this paper the subtraction algorithm is exercised through the mixed models.

The implementation, however, is not uniform, and the variation is itself informative: 27 accurate models let us separate what training
reliably converges on (the algorithm) from what it leaves free (binding and timing). Tab.~\ref{tab:FamilyInvariants} summarizes: the top
block holds in every accurate model; the bottom block varies without breaking any of them.

\begin{table*}[t]
    \caption{The family-level picture across the 29 accurate models: the algorithm is invariant, the implementation is not. Top block: properties that hold in every accurate model analyzed. Bottom block: properties that vary across models without breaking the algorithm. ``Family'' = all accurate models via the released per-model maps; ``worked examples'' = the deep-dive models of Sec.~\ref{sec:CausalSV}.}
    \label{tab:FamilyInvariants}
    \vspace{0.1in}
    \centering
    \small
    \begin{tabularx}{\textwidth}{|>{\raggedright\arraybackslash}X|p{5.5cm}|}
        \hline
        \textbf{Invariant across every accurate model} & \textbf{Evidence (scale)} \\
        \hline
        Every required sub-task role (\SAC \STC \SV; plus $MD$, $ND$, $MB$, \OPR, \SGN in mixed models) is present at a valid position, and the ordering constraints hold & node search + ablation + constraint checks (family) \\
        \hline
        The per-digit carry is computed eagerly, in place, as a binary make-carry & activation patching (worked examples) \\
        \hline
        The resolved carry is delivered carry-specifically to an answer-position combiner by redundant heads; ``='' is a depot, not the source & activation + edge path-patching with matched controls (worked examples) \\
        \hline
        The combiner \SV is a step function & interpolation of the combiner input (5 addition sizes + mixed) \\
        \hline
        No single node is necessary; necessity appears at the class level & ablation vs.\ untagged baseline ($d{=}5$--$13$) \\
        \hline 
        \textbf{Varies across models --- without breaking the algorithm} & \\
        \hline
        Which attention head implements a given sub-task; some sub-tasks split across two co-located heads & family maps \\
        \hline
        Degree of redundancy; the class-vs-single-node necessity gap \emph{grows} with digit count---intrinsic, not small-model slack & ablation ($d{=}5$--$13$) \\
        \hline
        Some models also compute the logically-redundant \SCC others omit it & family maps \\
        \hline
        PCA cluster geometry per head --- always the same three \ST classes (Fig.~\ref{fig:PCATrigrams5D}) & PCA (worked examples) \\
        \hline
        Timing: the resolved carry appears at the answer read, with a lag growing with size (2 tokens at 5--6 digits; 3 at 8) & linear probes ($d{=}5$--$8$) \\
        \hline
        Which route (residual stream vs.\ last-layer attention) delivers the mixed model's carry/borrow, per class and cascade depth & patching with deciding-matched nulls (6d + 8d mixed) \\
        \hline
        Mixed borrow delivery: spread across three heads at 6 digits, concentrated into one at 8 & attention profile, OV decode, group interchange (6d + 8d mixed) \\
        \hline
        At large $n$ the carry-axis separation compresses---consistent with dynamic-range compression of a place-value code, not an algorithm change & linear probes ($d{=}10$--$13$) \\
        \hline
    \end{tabularx}
\end{table*}

A representative addition node-map is shown in Tab.~\ref{tab:MathsPurposePerNodeAdd6D} (App.~\ref{app:ExperimentalModels}); the released per-model maps cover every size.

The located skeleton is also \emph{sufficient}, not merely present. Keeping only the map's useful nodes and mean-ablating the position-specific complement 
retains 0.94 of baseline addition accuracy on a 5-digit model (48 of 152 nodes kept; harsher resample-ablation gives 0.62), and 0.89 / 0.92 for addition on 
the 6- and 8-digit mixed models---while keeping a \emph{random} same-size node set retains 0.00 under both ablations: the mapped circuit specifically carries 
the computation. The test also exposes the map's limits: mixed-model subtraction retains only ${\sim}0.5$ / ${\sim}0.37$ (positive-/negative-answer) at 6 digits 
and 0.19 / 0.66 at 8---a model-specific shortfall that Sec.~\ref{sec:MixedPoly} localizes and closes. (Single seed per model and class.)

\subsubsection{Causally confirming the resolved-carry}
\label{sec:CausalSV}

Locating the \SAC \SC and \ST nodes establishes the per-digit pieces, but the harder claim---that the model computes the \emph{multi-digit} resolved carry \SV, 
which must propagate across a run of 9s as in $555555555+444444448$---was not causally confirmed by prior work. We close this gap using activation patching,
edge path-patching, and a step-function interpolation of the combiner input (App.~\ref{app:Techniques}), on all four worked-example models.

Tracing the carry gives a consistent picture across all four models. Each digit's carry is computed eagerly, in place, as a binary make-carry; the multi-digit carry 
is then propagated and resolved, and integrated into the answer by a step function at the answer position. The delivery is carry-specific---an interchange that 
toggles only the deciding carry flips exactly the affected answer digit, while same-class controls do not---and redundant, with no single node necessary; the 
running carry is sourced from the question-tail carry nodes, not read from the ``='' token. This \SV interface replicates across five addition sizes
($d = 5, 6, 8, 10, 13$) and on the mixed models at both sizes (Sec.~\ref{sec:MixedPoly}).

The family also shows a graded variation that our per-model analysis is built to capture (Tab.~\ref{tab:FamilyInvariants}, bottom block): the carry is computed 
in place, but its multi-digit resolution is represented \emph{lazily}, at the answer read, and this lag grows modestly with size---a 2-token deferral at 5--6 digits and 3 tokens at 8 digits.

We are explicit about the limits: at larger digit-counts the carry \emph{source} is probe-limited (the robust cross-size evidence is the
step combiner, the redundancy, and the answer-side resolution locus), and we show the carry is \emph{delivered} and \emph{combined} without
establishing that any single head \emph{selects} the deciding digit (see Limitations).

\subsubsection{Mixed models and polysemantic reuse}
\label{sec:MixedPoly}

Our accurate mixed (addition-and-subtraction) models contain the same skeleton, and show the same variability, as the addition models. The \SV findings above replicate causally on both the 6- and 8-digit mixed models, for all three question classes (addition, positive- and negative-answer subtraction), with the three classes' step combiners co-located on the same last-layer answer-position MLPs---a shared combiner.
The mixed models additionally show how one network serves several operations. \citet{zhang2024interpretingimprovinglargelanguage} noted shared heads across addition and subtraction; we find the reuse is pervasive, and it arises whether or not the mixed model is initialized from an addition model.

In mixed models initialized from a smaller addition model (Sec.~\ref{sec:TrainingNDigitMixed}), the majority of inherited nodes become polysemantic during training: rather than growing new subtraction circuits, the model adapts its addition circuits to both operations. A node that originally performed only Base Add (\SA) typically learns three sub-tasks---addition (\SA), positive-answer subtraction (\MD), and negative-answer subtraction (\ND); the full per-node map is in App.~\ref{app:NDigitSubtraction}, Tab.~\ref{tab:MathsPurposePerNodeMix6D}. Nodes that handled carry (\ST) likewise adapt to borrow (\MT).

This reuse is natural---the three operations share structure, each mapping a digit pair to a single-digit output---and extensive: most inserted addition-model nodes serve two or three operations (Tab.~\ref{tab:Coverage_SNM}).

\begin{table}[h]
    \caption{Mixed models re-use most inserted addition-model nodes. Many inserted nodes become polysemantic during training - performing addition, positive-answer subtraction \textbf{and} negative-answer subtraction subtasks simultaneously. For a sample mixed model that uses 96 nodes and had 48 nodes inserted, this table shows inserted node reuse.}    
    \label{tab:Coverage_SNM}
    \vspace{0.1in}
    \centering
    \begin{tabularx}{\columnwidth}{|X|c c|c c|}
        \hline
        & \multicolumn{2}{c|}{\textbf{Used}} & \multicolumn{2}{c|}{\textbf{Inserted}} \\
        \hline
        \textbf{Question class} & \textbf{\#} & \textbf{\%} & \textbf{\#} & \textbf{\%} \\
        \hline
        All questions & 96 & & 48 & \\
        Addition & 61 & 64\% & 42 & 88\% \\
        Positive-answer subtraction & 70 & 73\% & 40 & 83\% \\
        Negative-answer subtraction & 53 & 55\% & 29 & 60\% \\
        \hline
    \end{tabularx}
\end{table}

Selection among these co-computed outputs is itself informative: patching the full pre-combiner selector state flips a subtraction answer to the correct addition digit (0.96)---the final-layer combiner is shared---yet no rank-1 ``operator direction'' achieves any flips: selection is a distributed, high-dimensional transformation, not a low-rank control signal. The carried \emph{data}, by contrast, is low-dimensional per operation family: the two subtraction classes read their resolved borrow along one shared direction ($|\cos|\approx0.9$, nuisance-controlled and stable across read digits), while the addition carry axis is near-orthogonal (${\sim}0.21$). Low-dimensional data, high-dimensional control (single model and seed).

Finally, the sufficiency test of Sec.~\ref{sec:SkeletonFamily} exposes---and closes---a real incompleteness in the ablation-built map. Keeping only map nodes preserves operation selection perfectly (sign token retained at 1.00, all classes) but loses specific subtraction answer digits. The missing mass is last-layer attention heads at the failing digits' answer positions: heads tagged useful \emph{elsewhere}, under-tagged here because each is individually redundant. Restoring roughly ten such nodes recovers subtraction to ${\geq}0.94$; a same-size random augmentation stays near the unrestored level until ${\sim}80$ nodes. These heads deliver the borrow: they attend to the lower-digit operands, their group write decodes the resolved borrow-in at 1.00, and patching the group flips the produced digit at 1.00 (deciding-matched null 0.00) while any single head does almost nothing---redundancy defeats per-node attribution. The pattern replicates on the 8-digit mixed model with one instructive difference: the delivery concentrates into a single head (reading the ``=''/sign staging region) rather than three---redundancy concentrates as size grows. (Single seed per model and class; App.~\ref{app:Techniques}.)

\section{Conclusion}
\label{sec:Conclusion}
We set out to establish, with calibrated confidence, the algorithm a family of low-loss transformers implements for integer addition and subtraction. Training 46 small models on edge-case-enriched data yields 29 exceeding 99.999\% accuracy on tasks of up to 13 digits. From their internals we derived novel, exact, left-to-right algorithms built on progressive refinement of carry and borrow uncertainty, then causally confirmed---via patching with matched controls on four worked examples---that the models implement them, closing the open multi-digit resolved-carry gap. Across the accurate family the algorithm is the invariant: every model carries the same circuit skeleton, while head assignment, redundancy, timing, and delivery route vary without breaking it---and the located circuits are sufficient, not merely present. Mixed models reuse addition circuitry polysemantically for subtraction, whether or not initialized from an addition model. We release the models, per-model maps, and toolkit.

\section*{Limitations}
\label{sec:LimitationsAndFutureWork}

\textbf{Evidence depth and scope.} Our deep causal battery ran on four worked-example models; for the rest of the accurate family the skeleton claim rests on automated map-level search and ablation, and map depth decreases with model size. At large digit counts the carry direction is weakly separated in activation space, so the \emph{source} of the running carry is probe-limited there. We show the resolved carry is delivered and combined, but do not establish that any single head \emph{selects} the deciding digit; much of the representational evidence is linear-probe based, on 2--3-layer models; and the sufficiency and restoration results are single-seed per model and class.

\textbf{Attribution method.} Our ablation-built maps have a demonstrated failure mode: nodes that matter only as a redundant group are under-tagged. We detected this by sufficiency masking and repaired it by group restoration, but where the borrow is \emph{resolved} upstream of its last-layer delivery remains unpinned, as do the residual-stream representations of OPR and SGN beyond their functional roles.

\textbf{External validity and future work.} These are ${\sim}$10M-parameter specialized models; frontier LLMs implement addition through different, hybrid circuits \citep{lindsey2025biology}, and we make no transfer claim. Our declarative sub-task language and constraint-testing framework needs refinement and calibrated certainty metrics, though parts have already been reused elsewhere \citep{harrasse2025tinysqlprogressivetexttosqldataset}. The released family---same algorithm, varying implementation---is a ready-made testbed for universality studies across seeds, scales, and operations.

\section*{Ethics Statement}
\label{app:EthicsStatement}
Our work aims to explain the inner workings of transformer-based language models, which may have broad implications for a wide range of applications. A deeper understanding of generative AI has dual usage. While the potential for misuse exists, we discourage it. The knowledge gained can be harnessed to safeguard systems, ensuring they operate as intended. It is our sincere hope that this research will be directed toward the greater good, enriching our society and preventing detrimental effects. We encourage responsible use of AI, aligning with ethical guidelines.

\bibliography{paper}

\appendix

\section{Appendix: Terminology and Abbreviations}
\label{app:Terminology}
These terms and abbreviations are used in this paper and the associated Colabs and python code:

\begin{itemize}
\item \textbf{Pn} : Model (input or output) token position. Zero-based. e.g. \textbf{P}18, \textbf{P}18L1H0
\item \textbf{Ln} : Model layer n. Zero-based. e.g. P18\textbf{L}1H2
\item \textbf{Hn} : Attention head n. Zero-based. e.g. P18L1\textbf{H}2
\item \textbf{Mn} : MLP neuron n. Zero-based
\item \textbf{PnLnHn} : Location / name of a single attention head, at a specified layer, at a specific token position
\item \textbf{PnLnMn} : Location / name of a single MLP neuron, at a specified layer, at a specific token position
\item \textbf{D} : First number of the pair question numbers
\item \textbf{Dn} : nth numeric token in the first question number. Zero-based. D0 is the units value
\item \textbf{D'} : Second number of the pair question numbers
\item \textbf{D'n} : nth token in the second question number. Zero-based. D0 is the units value
\item \textbf{A} : Answer to the question (including answer sign)
\item \textbf{An} : nth token in the answer. Zero-based. A0 is the units value. The highest token is the “+" or “-" answer sign
\item \textbf{S} : Prefix for Addition. Think S for Sum. 
\item \textbf{SA} : Base Add. An addition subtask. $SA_n$ is defined as (Dn + D'n) \% 10. e.g. 5 + 7 gives 2
\item \textbf{SC} : Carry One. An addition subtask. $SC_n$ is defined as Dn + D'n $>=$ 10. e.g. 5 + 7 gives True
\item \textbf{ST} : TriCase. An addition subtask. Refer paper body for details
\item \textbf{M} : Prefix for Subtraction with a positive answer. Think M for Minus. Aka SUB
\item \textbf{MD}: Basic Difference. A subtraction subtask. $MD_n$ is defined as (Dn - D'n) \% 10. e.g. 3 - 7 gives 6
\item \textbf{MB}: Borrow One. A positive-answer subtraction subtask. $MB_n$ is defined as Dn - D'n $<$ 0. e.g. 5 - 7 gives True
\item \textbf{Node} : A logical locus implementing a specific subtask. Physically an attention head (e.g. P18L0H1), an MLP layer (P18L0MLP), or occasionally two heads working together.
\item \textbf{N} : Prefix for Subtraction with a negative answer. Think N for Negative. Aka NEG
\item \textbf{ND} : Basic Difference. A negative-answer subtraction subtask. $ND_n$ is defined as (Dn - D'n) \% 10. e.g. 3 - 7 gives 6
\item \textbf{NB} : Borrow One. A negative-answer subtraction subtask. $NB_n$ is defined as Dn - D'n $<$ 0. e.g. 5 - 7 gives True
\item \textbf{OPR} : Operator. A subtask that attends to the + or - token in the question (which determines whether the question is addition or subtraction).
\item \textbf{SGN} : Sign. A subtask that attends to the first answer token, which is + or -
\item \textbf{PCA} : Principal Component Analysis
\item \textbf{EVR} : Explained Variance Ratio. In PCA, EVR represents the percentage of variance explained by each of the selected components.
\end{itemize}

\section{Appendix: Training Data}
\label{app:TrainingData}
Each digit is represented as a separate token: \citet{liu2023goat} attribute LLaMA's ``remarkable arithmetic ability'' mainly to its consistent tokenization of numbers.

Training uses a new batch of data each step (aka Infinite Training Data) to minimize memorization. Depending on the configuration, each training run processes 1 to 4 million training datums. For the 5-digit addition problem there are 100,000 squared (that is 10 billion) possible questions. So the training data is much less than 1\% of the possible problems.

Addition and subtraction include rare edge cases. For example, these cascades (e.g. 55555+44446=100001, 54321+45679=1000000,
44450+55550=10000, 1234+8769=10003) are exceedingly rare: for 10-digit questions the deepest-cascade class occurs at frequency
${\sim}3\times10^{-4}$. The data generator was enhanced to increase the frequency of all known edge cases, lowering model loss.

We enriched 60\% of training data based on these edge cases (leaving the other 40\% of training data random):
(1) The initial model failed at cascading carry ones in addition. To rectify this, we randomly selected one operand and modified a random subset of digits in that operand to make the selected digit-position sum to 9, increasing the likelihood of a cascading carry one. 
(2) The initial model was worse at the subtraction task when the answer was negative, so we added 1 to each second operand digit (that was 8 or less) to increase the frequency of negative answers. 
(3) When the operands were identical, the initial model predicted -000000 instead of +000000. We increased the frequency of this case. For example, for 6 digit questions, we increased the frequency from 0.0001\% to 0.6\%.

\section{Appendix: Training Details}
\label{app:ModelShape}
\label{app:ModelLoss}

Training runs execute in a Colab notebook on a T4 GPU; each run takes up to 60 minutes.
The Colabs are available in the accompanying code repository.

The batch size is 64, with an AdamW optimizer (learning rate 0.00008, weight decay 0.1, betas (0.9, 0.98)); the learning rate warms up linearly over the first fifth of training and is then cosine-annealed. Per-digit loss is the negative log likelihood of an answer digit; all-digits loss is the mean across answer digits. Each model's final loss is in Tabs.~\ref{tab:AdditionModels} and ~\ref{tab:MixedModels}.

While we wanted very low loss, we also wanted to keep the model compact---intuiting that a smaller model is easier to understand. Things we tried to reduce loss that \textbf{didn't} work:

\begin{itemize}
\item Increasing the frequency of hard (cascading carry one) examples in the training data so the model has more hard examples to learn from. This improved training speed but did not reduce loss.
\item Increasing the number of attention heads from 3 to 4 or 5 (while still using 1 layer) to provide more computing power.
\item Changing the question format from “12345+22222=” to “12345+22222equals” giving the model more prediction steps after the question is revealed before it needs to state the first answer digit.
\item Inserting “+” in the answer format (e.g. “12345+22222=034567” becomes “12345+22222=+034567” had no impact on accuracy or the algorithm.
\item Changing the n\_layers to 2 and n\_heads to 2.
\end{itemize}

The smallest model shape that did reduce loss significantly was 2 layers with 3 attention heads.

\begin{figure*}[h]
\centering
\includegraphics[width=0.9\textwidth]{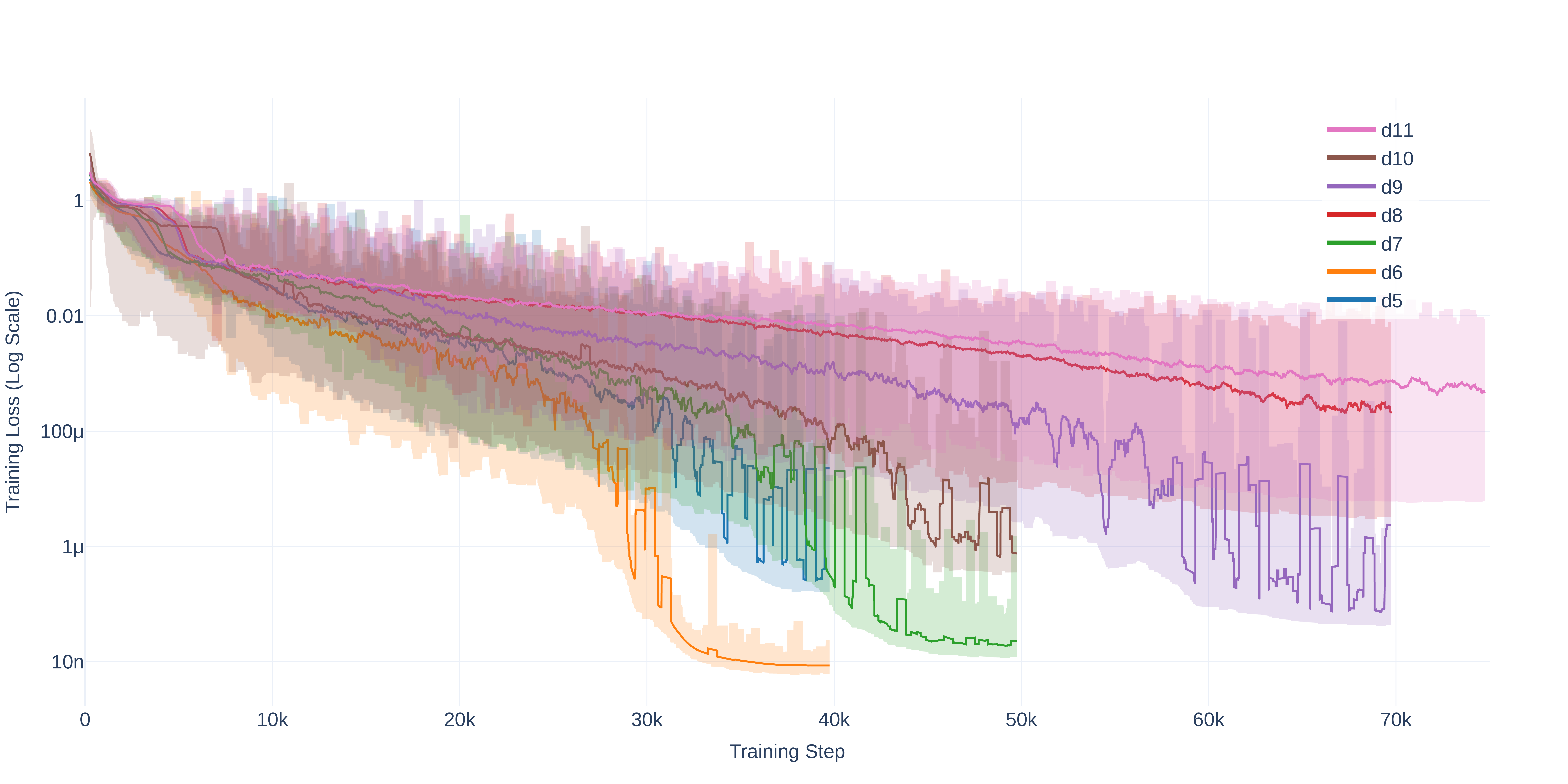}
\caption{The 5- to 8-digit initialized mixed models have very low loss and \textgreater{99.999}\% accuracy. With more digits the model loss and accuracy is worse. Details in Tab.\ref{tab:MixedModels}.}
\label{fig:TrainingLossMixed}
\end{figure*}

\section{Appendix: Analysis Techniques}
\label{app:Techniques}

We apply a common set of techniques to every model. Tab.~\ref{tab:Techniques} names each technique, what it tests, what it establishes, the models/scale it was applied at, and our confidence. The first group locates the circuit skeleton across the whole family; the second group causally confirms the resolved-carry (\SV) sub-task on the four worked-example models (6- and 8-digit addition and mixed); the third group tests circuit sufficiency and operation selection.

\begin{table*}[h]
    \caption{Analysis techniques applied in this work. ``Family'' = all accurate models; ``worked examples'' = 6d/8d addition and mixed; ``5 sizes'' = addition $d = 5, 6, 8, 10, 13$.}
    \label{tab:Techniques}
    \vspace{0.1in}
    \centering
    \small
    \renewcommand{\arraystretch}{1.15}
    \begin{tabularx}{\textwidth}{|
        >{\hsize=0.85\hsize\raggedright\arraybackslash}X|
        >{\hsize=1.30\hsize\raggedright\arraybackslash}X|
        >{\hsize=1.30\hsize\raggedright\arraybackslash}X|
        >{\hsize=0.95\hsize\raggedright\arraybackslash}X|
        >{\hsize=0.60\hsize\raggedright\arraybackslash}X|}
        \hline
        \textbf{Technique} & \textbf{What it tests} & \textbf{What it establishes} & \textbf{Scale} & \textbf{Conf.} \\
        \hline
        Node search & Does a node for a (sub-task, position) meet the attention/position/structure constraints? & The circuit skeleton is present & Family & High \\
        \hline
        Ablation & Is a node load-bearing for its sub-task? & Necessity vs.\ redundancy & Family & Med-High \\
        \hline
        PCA of node output & Does a node separate its sub-task classes (e.g.\ 3 $ST$ clusters)? & Representational class structure & Worked examples & Medium \\
        \hline
        Activation patching (interchange) & Does swapping a node's activation flip the answer as the algorithm predicts? & Causal role of make-carry / combiner nodes & Worked examples & Med-High \\
        \hline
        Edge path-patching & Does a specific node$\rightarrow$combiner edge carry the deciding carry? & Carry-specific, redundant delivery & Worked examples & Medium \\
        \hline
        Step-function interpolation & How does the combiner map its carry input to its output? & The combiner is a step function & 5 sizes + mixed & Med-High \\
        \hline
        Linear probes / cross-depth transfer & Where and when is the resolved carry linearly present? & Answer-read (L1) locus; lazy propagation & Worked examples & Medium \\
        \hline
        Sufficiency masking & Keeping only the map's nodes, does accuracy survive destroying the complement? & Map is sufficient and specific for addition; localizes the mixed subtraction gap & 5d addition + 6d/8d mixed & Med-High \\
        \hline
        Group restoration + OV decode + group interchange & What do the map-missed nodes compute, and does restoring them recover accuracy? & Redundant last-layer borrow-delivery heads; compact restore recovers subtraction & 6d + 8d mixed & Medium \\
        \hline
        Selector state-patch \& rank-1 steer & Is add-vs-sub selection a low-rank control signal? & Selection is distributed, high-dimensional; the combiner is shared & 6d mixed & Med-High \\
        \hline
    \end{tabularx}
\end{table*}

\section{Appendix: Experimental models}
\label{app:ExperimentalModels}

Forty-six models were trained and analyzed (Refer Tabs.~\ref{tab:AdditionModels} and ~\ref{tab:MixedModels}): 16 addition-only, 7 subtraction-only, and 23 mixed. The released collection is a superset of the models studied here: 48 model weight-sets, with per-model analysis repos (model.pth, training\_loss.json, behaviors.json, features.json) for 53 models, including 20-digit and training-variant models not analyzed in this paper.
The models and analysis output are available on Hugging Face to support further research in AI Safety.

\begin{table*}[h]
    \caption{For a sample 6-digit addition model, we show a compacted location map for the \SAC \SC and \ST subtasks. Interestingly 1) the \ST nodes are in a semi-random order 2) the second $ST1$ node is redundant 3) the model uses \SC nodes which are redundant 4) each \SA subtask is shared across two attention heads.}
    \label{tab:MathsPurposePerNodeAdd6D}
    \vspace{0.1in}
    \sffamily
    \centering
    \renewcommand{\arraystretch}{1.2}
    \begin{tabularx}{\textwidth}{|c|*{10}{>{\centering\arraybackslash}X|}}
    \hline 
    & \textcolor{brown}{\textbf{(P11)}} & \textcolor{brown}{\textbf{(P12)}} & \textcolor{brown}{\textbf{(P13)}} & \textcolor{brown}{\textbf{(P14)}} & \textcolor{brown}{\textbf{(P15)}} & \textcolor{brown}{\textbf{(P16)}} & \textcolor{brown}{\textbf{(P17)}} & \textcolor{brown}{\textbf{(P18)}} & \textcolor{brown}{\textbf{(P19)}} & \textcolor{brown}{\textbf{(P20)}} \\ 
    & \textcolor{brown}{\textbf{D'1}} & \textcolor{brown}{\textbf{D'0}} & \textcolor{brown}{\textbf{=}} & \textcolor{brown}{\textbf{+}} & \textcolor{brown}{\textbf{A6}} & \textcolor{brown}{\textbf{A5}} & \textcolor{brown}{\textbf{A4}} & \textcolor{brown}{\textbf{A3}} & \textcolor{brown}{\textbf{A2}} & \textcolor{brown}{\textbf{A1}} \\ 
    \hline 
    \textcolor{brown}{\textbf{L0H0}} & ST2 & ST3 & ST1 & ST4 & SC4 & SC3 & SC2 & SC1 & SC0 & \cellcolor{black!80}\\ 
    \hline
    \textcolor{brown}{\textbf{L0H1}} & ST1 & \cellcolor{black!80} & ST0 & \cellcolor{black!80} & \multirow{2}{*}{SA5} & \multirow{2}{*}{SA4} & \multirow{2}{*}{SA3} & \multirow{2}{*}{SA2} & \multirow{2}{*}{SA1} & \multirow{2}{*}{SA0} \\  
    \cline{1-1}
    \textcolor{brown}{\textbf{L0H2}} & \cellcolor{black!80} & \cellcolor{black!80} & \cellcolor{black!80} & ST5 & & & & & &\\ 
    \hline
    \end{tabularx}
\end{table*}

\begin{table*}[t]
  \caption{Addition-only models studied. The number of addition failures per million questions is shown. The number of useful attention heads at token position and useful MLP layers at token position are shown - summarizing the data shown in figures like Fig~\ref{fig:FailPerc5D}. }  
  \label{tab:AdditionModels}  
  \centering
  \begin{tabularx}{\textwidth}{|p{0.5cm}|p{0.5cm}|p{0.5cm}|p{0.7cm}|p{1cm}|p{1cm}|p{1.4cm}|p{0.8cm}|p{0.7cm}|X|}
    \hline
    \\[0.25em]  
    \textbf{\rotatebox{90}{\makebox[0pt]{Digits}}} & \textbf{\rotatebox{90}{\makebox[0pt]{Layers}}} & \textbf{\rotatebox{90}{\makebox[0pt]{Heads}}} & \textbf{Train Steps} & \textbf{Train Seed} & \textbf{Train loss} & \textbf{Add Fails / M} & \textbf{Heads used} & \textbf{MLPs used} & \textbf{Clopper-Pearson 95\%  Interval}\\ 
    \hline
    5 & 1 & 3 & 30K & 372001 & 9.4e-2 & 12621 & 15 & 6 & [1.24e-2, 1.28e-2] \\ 
    5 & 2 & 3 & 15K & 372001 & 1.6e-8 & 0 & 30 & 16 & [0.00e+0, 3.69e-6]\\
    5 & 2 & 3 & 40K & 372001 & 2.0e-9 & 0 & 22 & 15 & [0.00e+0, 3.69e-6]\\ 
    6 & 2 & 3 & 15K & 372001 & 1.7e-8 & 2 & 31 & 17 & [2.42e-7, 7.22e-6]\\
    6 & 2 & 3 & 20K & 173289 & 1.5e-8 & 0 & 28 & 17 & [0.00e+0, 3.69e-6]\\
    6 & 2 & 3 & 20K & 572091 & 7.0e-9 & 0 & 35 & 17 & [0.00e+0, 3.69e-6]\\
    6 & 2 & 3 & 40K & 372001 & 2.0e-9 & 0 & 29 & 17 & [0.00e+0, 3.69e-6]\\
    7 & 2 & 3 & 45K & 173289 & 3.0e-9 & 0 & 31 & 20 & [0.00e+0, 3.69e-6]\\
    8 & 2 & 3 & 45K & 173289 & 3.0e-9 & 0 & 35 & 22 & [0.00e+0, 3.69e-6]\\
    9 & 2 & 3 & 45K & 173289 & 3.0e-9 & 0 & 54 & 27 & [0.00e+0, 3.69e-6]\\
    10 & 2 & 3 & 40K & 572091 & 7.0e-9 & 0 & 44 & 28 & [0.00e+0, 3.69e-6]\\
    11 & 2 & 3 & 50K & 572091 & 8.0e-9 & 2 & 56 & 29 & [2.42e-7, 7.22e-6]\\
    12 & 2 & 3 & 50K & 572091 & 5.0e-9 & 3 & 50 & 33 & [4.25e-7, 7.88e-6]\\
    13 & 2 & 3 & 50K & 572091 & 6.3e-8 & 1 & 66 & 31 & [2.53e-8, 5.57e-6]\\
    14 & 2 & 3 & 60K & 572091 & 5.5e-6 & 199 & 68 & 35 & [1.74e-4, 2.00e-4] \\
    15 & 2 & 3 & 80K & 572091 & 8.6e-8 & 10 & 93 & 58 & [4.71e-6, 1.67e-5]\\    
    \hline
  \end{tabularx}   
\end{table*}

\begin{table*}[t]
  \caption{Subtraction-only and mixed models studied. The number of addition and subtraction failures per million questions is shown. The number of useful attention heads at token position and useful MLP layers at token position are shown - summarizing the data shown in figures like Fig~\ref{fig:FailPerc5D}. }  
  \label{tab:MixedModels}   
  \centering
  \begin{tabularx}{\textwidth}
{|p{0.3cm}|p{0.3cm}|p{0.3cm}|p{0.8cm}|p{0.9cm}|p{1cm}|p{1.4cm}|p{1.4cm}|p{0.8cm}|p{0.8cm}|X|}     
    \hline
    \\[0.25em]  
    \textbf{\rotatebox{90}{\makebox[0pt]{Digits}}} & \textbf{\rotatebox{90}{\makebox[0pt]{Layers}}} & \textbf{\rotatebox{90}{\makebox[0pt]{Heads}}} & \textbf{Train Steps} & \textbf{Train Seed} & \textbf{Train loss} & \textbf{Add Fails / M}& \textbf{Sub  Fails / M} & \textbf{Heads used} & \textbf{MLPs used} & \textbf{Clopper-Pearson 95\% Interval}\\ 
    \hline
    \multicolumn{11}{|c|}{\textcolor{lightbrown}{\textbf{Subtraction models}}} \\
    \hline
    5 & 2 & 3 & 30K & 372001 & 1.0e-3 & N/A & 3689 & 57 & 20 & [3.57e-3, 3.81e-3] \\
    6 & 2 & 3 & 30K & 372001 & 5.8e-6 & N/A & 2 & 37 & 21 & [2.42e-7, 7.22e-6] \\
    6 & 2 & 3 & 30K & 572091 & 5.8e-4 & N/A & 3889 & 58 & 21 & [3.77e-3, 4.01e-3] \\
    8 & 2 & 3 & 50K & 173289 & 4.0e-9 & N/A & 1 & 65 & 26 & [2.53e-8, 5.57e-6] \\
    8 & 2 & 3 & 50K & 371793 & 2.5e-5 & N/A & 487 & 71 & 28 & [4.45e-4, 5.32e-4] \\
    10 & 2 & 3 & 75K & 173289 & 2.0e-3 & N/A & 6672 & 101 & 37 & [6.50e-3, 6.72e-3] \\
    12 & 2 & 3 & 75K & 371793 & 3.4e-4 & N/A & 2175 & 96 & 32 & [2.08e-3, 2.25e-3] \\
    \hline
    \multicolumn{11}{|c|}{\textcolor{lightbrown}{\textbf{Mixed models}}} \\
    \hline
    5 & 3 & 4 & 40K & 372001 & 9.0e-9 & 0 & 0 & 45 & 21 & [0.00e+0, 3.69e-6]\\
    6 & 3 & 4 & 40K & 372001 & 5.0e-9 & 1 & 0 & 54 & 26 & [2.53e-8, 5.57e-6]\\
    7 & 3 & 4 & 50K & 372001 & 2.0e-8 & 2 & 6 & 108 & 40 & [3.45e-6, 1.58e-5]\\
    8 & 3 & 4 & 60K & 173289 & 4.7e-8 & 0 & 7 & 123 & 45 & [2.81e-6, 1.44e-5\\
    9 & 3 & 4 & 60K & 173289 & 3.2e-7 & 1 & 33 & 140 & 46 & [2.35e-5, 4.75e-5]\\
    10 & 3 & 4 & 75K & 173289 & 1.1e-6 & 2 & 295 & 143 & 53 & [2.65e-4, 3.53e-4]\\
    11 & 3 & 4 & 80K & 572091 & 3.9e-8 & 0 & 13 & 138 & 50 & [6.52e-6, 2.33e-5]\\
    12 & 3 & 4 & 85K & 572091 & 1.7e-8 & 2 & 10 & 167 & 55 & [8.68e-6, 2.72e-5]\\
    13 & 3 & 4 & 85K & 572091 & 9.5e-6 & 399 & 4164 & 197 & 64 & [4.41e-3, 4.70e-3]\\
    \hline
    \multicolumn{11}{|c|}{\textcolor{lightbrown}{\textbf{Mixed models initialized with addition model}}} \\
    \hline
    5 & 2 & 3 & 40K & 572091 & 2.5e-7 & 1 & 52 & 49 & 20 & [5.35e-5, 8.48e-5]\\
    6 & 2 & 3 & 40K & 572091 & 2.4e-8 & 0 & 5 & 57 & 21 & [1.63e-6, 1.10e-5]\\
    6 & 3 & 3 & 40K & 572091 & 1.8e-8 & 0 & 3 & 70 & 35 & [4.25e-7, 7.88e-6]\\
    6 & 3 & 3 & 80K & 572091 & 1.6e-8 & 0 & 3 & 75 & 35 & [4.25e-7, 7.88e-6]\\
    6 & 3 & 4 & 40K & 372001 & 8.0e-9 & 0 & 0 & 72 & 26 & [0.00e+0, 3.69e-6]\\
    6 & 3 & 4 & 40K & 173289 & 1.4e-8 & 3 & 2 & 60 & 29 & [2.16e-6, 1.18e-5]\\
    6 & 3 & 4 & 50K & 572091 & 2.9e-8 & 0 & 4 & 79 & 29 & [8.11e-7, 9.28e-6]\\
    7 & 3 & 4 & 50K & 572091 & 1.8e-8 & 4 & 1 & 104 & 38 & [1.03e-6, 1.08e-5]\\
    8 & 3 & 4 & 70K & 572091 & 4.3e-5 & 50 & 1196 & 116 & 42 & [1.11e-3, 1.23e-3]\\
    9 & 3 & 4 & 70K & 572091 & 5.4e-8 & 1 & 4 & 160 & 50 & [8.11e-7, 9.28e-6]\\
    10 & 3 & 3 & 50K & 572091 & 6.3e-7 & 6 & 7 & 90 & 45 & [5.64e-6, 2.15e-5]\\
    11 & 3 & 4 & 75K & 572091 & 5.9e-5 & 11066 & 1120 & 141 & 47 & [1.19e-2, 1.22e-2]\\
    \hline
    \multicolumn{11}{|c|}{\textcolor{lightbrown}{\textbf{Mixed models initialized with add model. Reset useful heads every 100 steps}}} \\
    \hline
    6 & 4 & 4 & 40K & 372001 & 1.7e-8 & 3 & 8 & 51 & 30 & [5.64e-6, 2.15e-5]\\
    \hline
    \multicolumn{11}{|c|}{\textcolor{lightbrown}{\textbf{Mixed models inited with add model. Reset useful hds \& MLPs every 100 steps}}} \\
    \hline
    6 & 4 & 3 & 40K & 372001 & 3.0e-4 & 17 & 3120 & 115 & 53 & [3.06e-3, 3.30e-3]\\
    \hline
  \end{tabularx}  
\end{table*}

For each model the training Colab notebook generates two files:
\begin{itemize}
\item A “model.pth" file containing the model weights
\item A “training\_loss.json" file containing configuration information and training loss data
\end{itemize}

While, for each model the analysis Colab notebook generates two more files:
\begin{itemize}
\item A “behaviors.json" file containing generic “behavior" facts learned about the model by the Colab e.g. P18L0H0 attends to tokens D3 and D'3
\item A “features.json" file containing maths-specific “feature" facts learned about the model by the Colab e.g. P18L0H0 performs the SC3 subtask.
\end{itemize}

\begin{table*}[h!]
    \caption{For a sample model, all nodes used in predictions are shown by token position (horizontally) and model layer (vertically), detailing the \textbf{answer digits} they impact. Here, the attention heads in token position P10 labeled A5..3 help predict the answer digits A3, A4 and A5. For all addition and mixed models studied, before the “=" token, each node often calculates data used to predict \textbf{multiple} answer digits. After the “=" token, all nodes in a given token position are used to predict a \textbf{single} answer digit.}
    \label{tab:Impact5D}
    \vspace{0.1in}
    \sffamily
    \centering
    \renewcommand{\arraystretch}{1.2}
    \begin{tabularx}{\textwidth}{|c|*{10}{>{\centering\arraybackslash}X|}}
    \hline 
    & \textcolor{brown}{\textbf{(P6)}} & \textcolor{brown}{\textbf{(P9)}} & \textcolor{brown}{\textbf{(P10)}} & \textcolor{brown}{\textbf{(P11)}} & \textcolor{brown}{\textbf{(P12)}} & \textcolor{brown}{\textbf{(P13)}} & \textcolor{brown}{\textbf{(P14)}} & \textcolor{brown}{\textbf{(P15)}} & \textcolor{brown}{\textbf{(P16)}} & \textcolor{brown}{\textbf{(P17)}} \\ 
    & \textcolor{brown}{\textbf{D'4}} & \textcolor{brown}{\textbf{D'1}} & \textcolor{brown}{\textbf{D'0}} & \textcolor{brown}{\textbf{=}} & \textcolor{brown}{\textbf{+}} & \textcolor{brown}{\textbf{A5}} & \textcolor{brown}{\textbf{A4}} & \textcolor{brown}{\textbf{A3}} & \textcolor{brown}{\textbf{A2}} & \textcolor{brown}{\textbf{A1}} \\ 
    \hline 
    \textcolor{brown}{\textbf{L0H0}} & \cellcolor{black!80} & \cellcolor{black!80} & & \cellcolor{black!80} & \cellcolor{black!80} & \multirow{2}{*}{A4} & \cellcolor{black!80} & \multirow{2}{*}{A2} & \multirow{2}{*}{A1} & \cellcolor{black!80}\\ 
    \cline{1-1} \cline{2-3} \cline{5-6} \cline{8-8} \cline{11-11} 
    \textcolor{brown}{\textbf{L0H1}} &  & A5 & A5..3 & \cellcolor{black!80} & \multirow{6}{*}{A5} &  & A3 &  &  & A0 \\  
    \cline{1-1} \cline{3-3} \cline{5-5} \cline{7-11}
    \textcolor{brown}{\textbf{L0H2}} & A5 & \cellcolor{black!80} &  & \multirow{2}{*}{A5..1} &  & \cellcolor{black!80} & \cellcolor{black!80} & \cellcolor{black!80} & \cellcolor{black!80} & \cellcolor{black!80}\\ 
    \cline{1-1} \cline{3-4} \cline{7-11}
    \textcolor{brown}{\textbf{L0MLP}} &  & \cellcolor{black!80} & A5..2 &  & & A4 & A3 & A2 & A1 & A0 \\ 
    \cline{1-5} \cline{7-11}
    \textcolor{brown}{\textbf{L1H0}} & \cellcolor{black!80} & \cellcolor{black!80} & \cellcolor{black!80} & \cellcolor{black!80} &  & \cellcolor{black!80} & \cellcolor{black!80} & \cellcolor{black!80} & \cellcolor{black!80} & \cellcolor{black!80}\\ 
    \cline{1-5} \cline{7-11} 
    \textcolor{brown}{\textbf{L1H2}} & \cellcolor{black!80} & \cellcolor{black!80} & \cellcolor{black!80} & \cellcolor{black!80} &  & \multirow{2}{*}{A4} & \multirow{2}{*}{A3} & \multirow{2}{*}{A2} & \cellcolor{black!80} & \cellcolor{black!80}\\ 
    \cline{1-5} \cline{10-11} 
    \textcolor{brown}{\textbf{L1MLP}} & \cellcolor{black!80} & \cellcolor{black!80} & \cellcolor{black!80} & \cellcolor{black!80} &  &  &  &  & A1 & A0 \\ 
    \hline
    \end{tabularx}
\end{table*}

\begin{figure}[ht]
\centering
\includegraphics[width=\columnwidth]{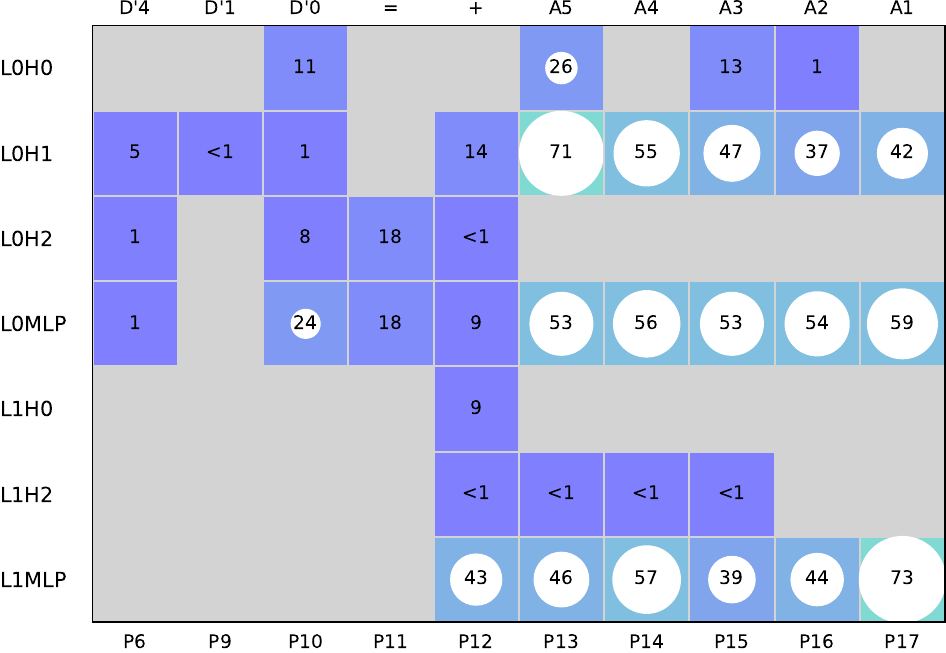}
\caption{This map shows the \textbf{\% of enriched questions} that fail when we ablate each node in a \textbf{5-digit} 2-layer 3-head addition model. The model only uses nodes in token positions P6 to P17 (i.e. tokens D'4 to A1). Lower percentages correspond to rarer edge cases. The gray space represents nodes that are not used by the model.}
\label{fig:FailPerc5D}
\end{figure}

\section{Appendix: Mixed Model Initialization}
\label{app:MixedModelInit}

We experimented with three ways to re-use the trained addition model in the mixed model: \textit{Initialize Only} (copy the addition weights into the untrained mixed model), \textit{Freeze Attention} (as before, but recopy the attention-head weights from the addition model every 100 training steps), and \textit{Freeze All} (recopy attention heads and MLP layers every 100 steps). Our intuition was that the freeze variants would keep the trained model easier to interpret, but only Initialize Only trained quickly to accuracy on both operations---the freeze variants constrain learning too much.

We also tested \emph{where} to insert the 2-layer 3-head addition model into the 3-layer 4-head mixed model (first vs.\ last layers; first vs.\ last heads). We initialized the first two layers and first three heads, reasoning that early placement makes reuse of the addition circuits---and hence interpretation---most likely.

\section{Appendix: Seed Sensitivity}
\label{app:SeedSensitivity}
\begin{figure*}[ht]
\centering
\includegraphics[width=0.9\textwidth]{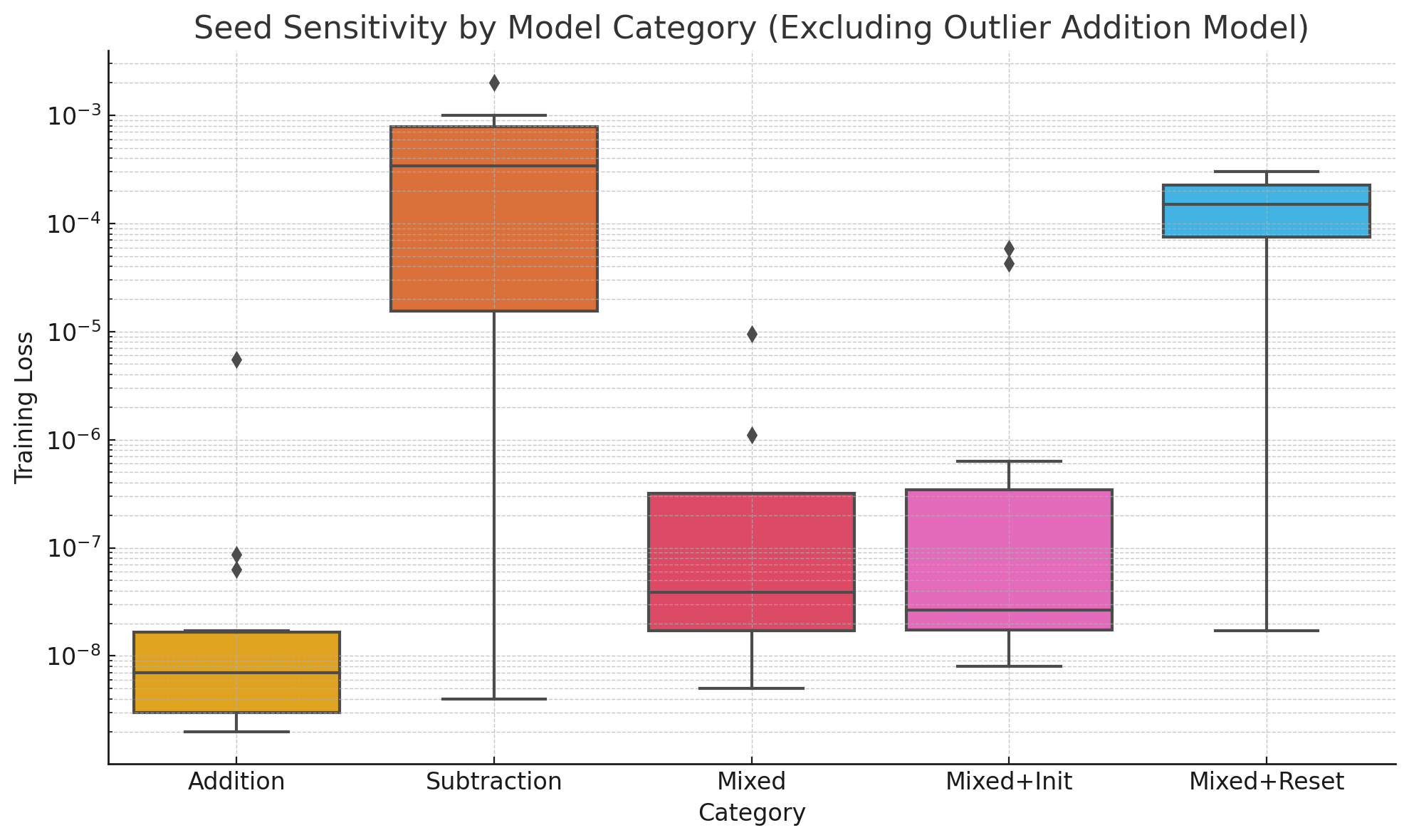}
\caption{A visualization of the range of training losses across 45 models grouped by the different model categories.}
\label{fig:SeedSensitivity}
\end{figure*}

An analysis of the sensitivity of 45 models to the initial seed was performed (This analysis excluded one model - the inaccurate 1-layer addition model that reproduces the \citet{quirke2023understanding} paper results). Fig.~\ref{fig:SeedSensitivity} shows the results. We conclude:

\begin{itemize}
\item The Addition models are the most stable - that is they are not sensitive to the seed value. 
\item The other categories (Subtraction, Mixed, Mixed+Init, and Mixed+Reset)  show relatively low to moderate sensitivity.
\item The higher average loss for Subtraction models show that the models find it harder to learn Subtraction is isolation.
\item The higher average loss for Mixed+Reset models show that the this type of intervention during training makes it harder for the models to learn 
\end{itemize}

\section{Appendix: N-Digit Subtraction}
\label{app:NDigitSubtraction}

The mixed models perform addition and subtraction accurately. Tab.~\ref{tab:MathsPurposePerNodeMix6D} gives the full per-node sub-task map, over the last five token positions, for the 6-digit mixed capstone model discussed in Sec.~\ref{sec:MixedPoly}.

\begin{table*}[h]
    \caption{For this mixed model, in the last 5 tokens, polysemantic attention heads simultaneously generate outputs for the three question classes (S, M and N). Other heads calculate the question class by attending to the question operation (OPR) token and the answer sign (SGN) token. The class-appropriate output is then selected via a distributed transformation feeding a shared last-layer combiner (Sec.~\ref{sec:MixedPoly}).}
    \label{tab:MathsPurposePerNodeMix6D}
    \vspace{0.1in}
    \sffamily
    \small
    \centering
    \renewcommand{\arraystretch}{1.3}

    \begin{tabularx}{\textwidth}{|c|*{12}{>{\centering\arraybackslash}X|}}

    \hline
    & \textcolor{brown}{\textbf{(P9)}} & \textcolor{brown}{\textbf{(P10)}} & \textcolor{brown}{\textbf{(P11)}} & \textcolor{brown}{\textbf{(P12)}} & \textcolor{brown}{\textbf{(P13)}} & \textcolor{brown}{\textbf{(P14)}} & \textcolor{brown}{\textbf{(P15)}} & \textcolor{brown}{\textbf{(P16)}} & \textcolor{brown}{\textbf{(P17)}} & \textcolor{brown}{\textbf{(P18)}} & \textcolor{brown}{\textbf{(P19)}} & \textcolor{brown}{\textbf{(P20)}} \\
    & \textcolor{brown}{\textbf{D'3}} & \textcolor{brown}{\textbf{D'2}} & \textcolor{brown}{\textbf{D'1}} & \textcolor{brown}{\textbf{D'0}} & \textcolor{brown}{\textbf{=}} & \textcolor{brown}{\textbf{A7}} & \textcolor{brown}{\textbf{A6}} & \textcolor{brown}{\textbf{A5}} & \textcolor{brown}{\textbf{A4}} & \textcolor{brown}{\textbf{A3}} & \textcolor{brown}{\textbf{A2}} & \textcolor{brown}{\textbf{A1}} \\
    \hline
    \multirow{3}{*}{\textcolor{brown}{\textbf{L0H0}}} & MT4 & \cellcolor{black!80} & \cellcolor{black!80}  & MT3 & \cellcolor{black!80}  & ST4 &  &  & SC2 & SC1 & SC0 & OPR \\
    & & \cellcolor{black!80} &  \cellcolor{black!80} & MT3 &  \cellcolor{black!80} & MT4 & SC4 & SC3 & NB2 & MB1 & MB0 & SGN\\
    & & \cellcolor{black!80} & \cellcolor{black!80} & & \cellcolor{black!80} & OPR & &  &  & NB1 & NB0 & \\ \cline{1-8}
    \hline

    \multirow{2}{*}{\textcolor{brown}{\textbf{L0H1}}} & ST4 & ST2 & ST1 & ST3 & ST0 & \cellcolor{black!80} & SA5 &  &  &  &  & \\
    & & MT2 & MT1 & & MT0 & \cellcolor{black!80} & MD5 & SA4 & SA3 & SA2 & SA1 & SA0\\ \cline{1-2}

    \multirow{3}{*}{\textcolor{brown}{\textbf{L0H2}}} & \cellcolor{black!80} & \cellcolor{black!80} & \cellcolor{black!80} & \cellcolor{black!80} & \cellcolor{black!80} & ST5 & SA4 & MD4 & MD3 & MD2 & MD1 & MD0\\
    & \cellcolor{black!80} & \cellcolor{black!80} & \cellcolor{black!80} & \cellcolor{black!80} & \cellcolor{black!80} & OPR & ND5 & ND4 & ND3 & ND2 & ND1 & ND0 \\
    & \cellcolor{black!80} & \cellcolor{black!80} & \cellcolor{black!80} & \cellcolor{black!80} & \cellcolor{black!80} & SGN & &  &  &  &  & \\ \hline

    \multirow{2}{*}{\textcolor{brown}{\textbf{L0H3}}} & \cellcolor{black!80} & \cellcolor{black!80} & \cellcolor{black!80} & \cellcolor{black!80} & \cellcolor{black!80} & \cellcolor{black!80} & OPR & OPR & OPR & OPR & OPR & OPR\\
    & \cellcolor{black!80} & \cellcolor{black!80} & \cellcolor{black!80} & \cellcolor{black!80} & \cellcolor{black!80} & \cellcolor{black!80} & SGN & SGN & SGN & SGN & SGN & SGN\\
    \hline
    \end{tabularx}
\end{table*}

One mixed-model sub-task stands out: the algorithm relies on calculations done at token position P0, when the model has seen only one question token. What can the model gather from just the first token? Intuitively, if that token is an “8" or “9" the answer sign is more likely to be “+" than “-". The model uses this heuristic even though the probabilistic information is sometimes incorrect and so works against achieving very low loss.

\section{Appendix: Addition Hypothesis 3}
\label{app:Hypothesis3}

The hypothesis 3 pseudo-code (Tab.~\ref{tab:Hypothesis3A5A4}) was derived iteratively by obtaining experimental results and mapping them to mathematical operations. To validate the underlying framework we implemented the algorithm in Python and executed one million additions with zero errors.

\begin{table}[ht]
    \centering
    \caption{Hypothesis 3 pseudo-code for the 5-digit addition worked example, restored from the published version in this paper's notation: each intermediate value, how it is computed, and the nodes implementing it. $ST_n = \mathrm{TriCase}(D_n, D'_n)$. The $SV_1$ recomputed at P14 defers to P10's value only when the two disagree (final bullet below).}
    \label{tab:Hypothesis3A5A4}
    \vspace{0.1in}
    \small
    \begin{tabularx}{\columnwidth}{p{0.7cm} p{3.6cm} p{2.6cm}}
         \textbf{Value} & \textbf{Calculated as} & \textbf{Nodes used} \\
         \hline
         $ST_2$ & TriCase($D_2$, $D'_2$) & P8.L0.H1 + MLP \\
         $ST_1$ & TriCase($D_1$, $D'_1$) & P9.L0.H1 + MLP \\
         $SV_1$ & TriAdd($ST_1$, $ST_0$) & P10.L0.H1 + MLP \\
         $SV_3$ & TriAdd($ST_3$, TriAdd($ST_2$, $SV_1$)) & P11.L0.H1 \\
         $ST_4$ & TriCase($D_4$, $D'_4$) & P11.L0.H2 \\
         $SV_4$ & TriAdd($ST_4$, $SV_3$) & P11.L0.MLP \\
         \textbf{A5} & $SV_4$ & P11.L1.MLP \\
         \hline
         $SA_4$ & $(D_4 + D'_4) \bmod 10$ & P12.L0.H0+H2 \\
         $SV_3$ & as above & P12.L0.H1 \\
         \textbf{A4} & $(SA_4 + SV_3) \bmod 10$ & P12 MLPs \\
         \hline
         $SA_3$ & $(D_3 + D'_3) \bmod 10$ & P13.L0.H0+H2 \\
         $SV_2$ & TriAdd($ST_2$, $SV_1$) & P13.L0.H1 \\
         \textbf{A3} & $(SA_3 + SV_2) \bmod 10$ & P13 MLPs \\
         \hline
         $SA_2$ & $(D_2 + D'_2) \bmod 10$ & P14.L0.H0+H2 \\
         $SV_1$ & recomputed; defers to P10 $SV_1$ & P14.L0.H1 \\
         \textbf{A2} & $(SA_2 + SV_1) \bmod 10$ & P14 MLPs \\
         \hline
         $SA_1$ & $(D_1 + D'_1) \bmod 10$ & P15.L0.H0+H2 \\
         $SC_0$ & $(D_0 + D'_0) \geq 10$ & P15.L0.H1 \\
         \textbf{A1} & $(SA_1 + SC_0) \bmod 10$ & P15 MLPs \\
         \hline
         \textbf{A0} & $(D_0 + D'_0) \bmod 10$ & P16.L0.H0+H2, MLPs \\
         \hline
    \end{tabularx}
\end{table}

Some of the experiments and mappings behind this pseudo-code were:

\begin{itemize}
\item Ablation experiments show that the A5 value is \textbf{accurately} calculated in prediction step 11 using 5 attention heads and 5 MLP layers. The pseudo-code accurately calculates A5 while constraining itself to this many steps.

\item Ablating the nodes one by one shows which answer digit(s) are reliant on each node (Ref Tab.~\ref{tab:Impact5D}). Most interestingly, ablating P10.L0.H1 impacts the answer digits A5, A4, A3, A2 (but not A1 and A0). This node is used in the calculation of A5, A4, A3, A2 in prediction steps 11, 12, 13 and 14. These relationships are constraints that are all obeyed by the pseudo-code. 

\item The pseudo-code has 4 instances where \ANST is calculated using TriCase. PCA of the corresponding nodes (P8.L0.H1, P9.L0.H1, P11.L0.H2 and P14.L0.H1) shows tri-state output for the specified \DN . (see Figure~\ref{fig:PCATrigrams5D}).

\item The pseudo-code has 4 instances where compound functions using TriCase and TriAdd to generate tri-state outputs. PCA of the corresponding nodes (P11.L0.H1, P12.L0.H1 and P13.L0.H1) shows tri-state output for the specified \DN . (see Figure~\ref{fig:PCATrigrams5D}).

\item Activation patching (aka interchange intervention) experiments at attention head level confirmed some aspects of the calculations.

\item The pseudo code includes calculations like ST1 which it says is calculated in P9.L0.H1 \textbf{and} P9.L0.MLP. Ablation tells us both nodes are necessary. For the attention head we use the PCA results for insights. We didn't implement a similar investigative tool for the MLP layer, so in the pseudo-code we attribute the calculation of ST1 to both nodes.

\item For P10.L0.H1, the attention head PCA could represent either a bi-state or tri-state output. The MLP layer at P10.L0.MLP could map the attention head output to either a bi-state or tri-state. We cannot see which. The pseudo-code shows a tri-state calculation at P10.L0.MLP, but with small alterations the pseudo-code would work with a bi-state output.

\item For P15.L0.H1 the attention head PCA could represent either a bi-state or tri-state output. The pseudo-code shows a bi-state calculation SC0 at P15.L0.H1, but with small alterations the pseudo-code would work with a tri-state output.

\item The calculation of ST2 in P14.L0.H1 is a interesting case. The model needs ST2 for A2 accuracy. The model could simply reuse the accurate ST2 value calculated in P10. Activation patching shows that it does not. Instead the P14 attention heads calculate ST1 from D1 and D'1 directly, and only relies on the P10.D1.ST2 value in the case where ST2 != ST1. That is, the calculation is “use P14.ST1 value else use ST2 value". This aligns with the model learning the P10.ST1 calculation early in training (for 90\% accuracy) and later learning that P10.ST2 contains additional information it can use to get to \textgreater{99.999}\% accuracy.
\end{itemize}

\end{document}